# AUTONOMOUS MOSQUITO HABITAT DETECTION USING SATELLITE IMAGERY AND CONVOLUTIONAL NEURAL NETWORKS FOR DISEASE RISK MAPPING

SRIRAM ELANGO [1], NANDINI RAMACHANDRAN [1], RUSSANNE LOW [2]

[1] Center for Space Research, University of Texas at Austin
[2] Institute for Global Environmental Strategies (IGES)

*Abstract* — Mosquitoes are known vectors for disease transmission that cause over one million deaths globally each year. The majority of natural mosquito habitats are areas containing standing water such as ponds, lakes, and marshes. These habitats are challenging to detect using conventional ground-based technology on a macro scale. Contemporary approaches, such as drones, UAVs, and other aerial imaging technology are costly when implemented. Multispectral imaging technology such as Lidar is most accurate on a finer spatial scale whereas the proposed convolutional neural network(CNN) approach can be applied for disease risk mapping and further guide preventative efforts on a more global scale. By assessing the performance of autonomous mosquito habitat detection technology, the transmission of mosquito-borne diseases can be prevented in a cost-effective manner. This approach aims to identify the spatiotemporal distribution of mosquito habitats in extensive areas that are difficult to survey using ground-based technology by employing computer vision on satellite imagery for proof of concept. The research presents an evaluation and the results of 3 different CNN models to determine their accuracy of predicting large-scale mosquito habitats. For this approach, a dataset was constructed utilizing Google Earth satellite imagery containing a variety of geographical features in residential neighborhoods as well as cities across the world. Larger land cover variables such as ponds/lakes, inlets, and rivers were utilized to classify mosquito habitats while minute sites such as puddles, footprints, and additional human-produced mosquito habitats were omitted for higher accuracy on a larger scale. Using the dataset, multiple CNN networks were trained and evaluated for accuracy of habitat prediction. Utilizing a CNN-based approach on readily available satellite imagery is cost-effective and scalable, unlike most aerial imaging technology. Testing revealed that YOLOv4 obtained greater accuracy in mosquito habitat detection than YOLOR or YOLOv5 for identifying large-scale mosquito habitats. YOLOv4 is found to be a viable method for global mosquito habitat detection and surveillance.

## 1. INTRODUCTION

Mosquitoes hold the leading cause of animal-related deaths in humans with over 725,000 deaths per year (ISGlobal, 2017). To help guide frontline efforts on disease prevention, it is therefore imperative to have a responsive and scalable solution for the detection of hotspot regions of such habitats. Understanding relationships between land cover components and mosquito population distribution is essential to predicting the likelihood of mosquito habitats to exist in certain regions. Recognizing the ecological conditions for suitable habitats, vector competence, and larvae development is necessary in order to classify what constitutes a large-scale mosquito habitat. For malaria, in particular, the geographical distribution is seasonally driven by the availability and conditions of water bodies, contributing to the need for constant mosquito habitat surveillance. The preferred breeding grounds for mosquito oviposition also vary among different types of species (Stanton *et al*, 2021). For example, species such as Culex pipiens and Culex quinquefasciatus tend to prefer stagnant, polluted, and smaller sources like puddles, gutters, and ponds. They carry a variety of diseases like West Nile Virus and St. Louis encephalitis. Culex mosquitoes thrive in human households so measuring land use and monitoring human-environment interactions play a large role in determining the presence of such mosquitoes. On the other hand, two main sibling species like the Anopheles gambiae and Anopheles arabiensis tend to be found in the edges of sunlit pools/rivers. This suggests that the breeding ground may vary depending on the type of mosquito species, though all rely on stagnant water. Applying the knowledge of breeding preferences, satellite imagery can be used to detect placement of different larvae species in mosquito habitats. This can aid in mapping out potential hotspot locations of various mosquito-borne vector diseases.

Classic survey techniques for autonomous mosquito habitat detection include aerial imaging technology like drones (Stanton, 2020) and UAVs (Dias *et al*, 2018) (Schenkel *et al*, 2020) as well as multispectral imaging technology such as Lidar (ARSC, 2020). Aerial technology is best for habitat detection on a micro-geographical scale. A scalable solution would be both unrealistic and expensive for employing drones across multiple countries as well as in under-resourced/inaccessible areas. Two recent studies use UAVs to correlate land use with Anopheles gambiae breeding sites in Thailand (Kaewwaen and Bhumiratana,

2015). However, there have been no such approaches targeting residential areas on a global scale. Another autonomous approach, Lidar detection, analyzes water bodies through spectral signatures for the presence of different mosquito larvae. One such study used Lidar and RGB mapping of 16 randomly sampled water bodies in the Loreto Region of the Peruvian Amazon Forest (Carrasco *et al*, 2019). This approach was conclusive in its classification of mosquito breeding sites but time-intensive, expensive, and confined to a smaller spatial scale. Due to the abundance of readily available satellite imagery, it is more effective to use CNNs and computer vision technology for object detection on mosquito habitats. A CNN approach is scalable and efficient. This makes it most equipped for dealing with large-scale priorities like disease risk mapping. This research presents an evaluation of different CNN models that were trained in the presence of a variety of land cover elements to determine its accuracy in predicting large-scale mosquito habitats. The proposed model is trained with readily available satellite imagery of mosquito habitats that are primarily located in close proximity to residential neighborhoods, although it can also be applied to any geographical setting. This information can aid in where to guide preventative measures for mosquito-borne vector disease control, label mosquito-spread patterns, and map low-resource areas, further potentially preventing millions of deaths to mosquito-borne diseases.

2. UTILIZATION OF GOOGLE EARTH SATELLITE IMAGERY

To prepare the training dataset, this study employed the use of cloud-based computing services like Google Earth Engine to capture satellite imagery data. Google Earth satellite imagery is a free, open-source alternative to purchasing high-resolution satellite imagery. 507 images were acquired from a variety of geographical regions across the world to include diverse topographical features. The images were then screen captured at an altitude of 1,300 meters[1] to facilitate imagery analysis using annotation software. Each image was annotated for object detection using bounding boxes via the Computer Vision Annotation Tool (CVAT). Three different classes were used to classify the mosquito habitats on the satellite imagery: Ponds/Lakes, River Inlets, and Rivers. This study centered on large-scale mosquito habitats, so a majority of the satellite imagery was taken in presence of these particular land cover elements. After annotation, the images were imported into Roboflow (Nelson and Dwyer, 2019) where they were compiled in one central dataset. The dataset was then divided using a 80 - 20 percent train/validate split and exported as a TXT file in both the Darknet and PyTorch framework based on the model being tested. The images were then accessed using a Google Colab Notebook where training configuration settings were adjusted to fit the model and dataset. Each model was tested on a set of 125 unannotated Google Earth satellite images to gauge its accuracy in prediction/inferences.

3. UTILIZATION OF CONVOLUTIONAL NEURAL NETWORKS

Convolutional Neural Networks (CNNs) are a form of artificial neural networks often utilized to process visual imagery, proposed by Fukushima (1980) and innovated by LeCun *et al* (1998) and Alex Krizhevsky *et al* (2017). The network receives an input image and assigns importance (learnable weights and biases) to differentiate features and characteristics in one image from another (Chowdhry *et al* 2021). Mimicking neurons within the human brain, the network's artificial neurons identify and weigh different components of the image inputted. Once trained, the CNN is further able to classify and identify components through its internal and custom adjusted weights. Currently employed and applied for a variety of uses, such as cancer detection and medicine, natural language processing, facial recognition, and more, CNNs can be further utilized in autonomous mosquito habitat remote sensing and classification. We apply convolutional neural networks due to the scalability and efficiency of such an approach, enabling an effective solution to mosquito habitat detection on a global scale.

*3.1. Comparison Against Common Classification Approaches*

Contemporary approaches in mosquito habitat classification often utilize drones and UAVs for finer spatial analysis and identification, however, they prove expensive and inefficient for immediate identification of mosquito habitats, a factor required in the circumstance of epidemics. Though a finer analysis is required for thorough identification of mosquito habitats, employing drones to map extensive areas and analyzing the data obtained necessitates a substantial quantity of time, insufficient for immediate response and action. To target and identify areas containing large-scale mosquito habitats in a timely and cost-efficient manner, CNNs prove beneficial as there is no current approach that effectively resolves this problem.

Satellite data additionally provides a finer analysis through the utilization of various precise sensors (Wimberly *et al*, 2021), however the analysis and identification of mosquito habitats through this approach demands an increased amount of time compared to the timely approach of a CNN. A CNN can be further applied to scan vast areas and thus identify regions that are vulnerable to mosquito disease transmission, map mosquito habitat distribution, and identify what pathways the diseases can spread based upon the spatial distribution of mosquito habitats.

Through this approach, authorities/government officials can use CNNs to immediately identify large-scale mosquito habitats that need to be targeted and addressed. As these habitats are being treated, a finer spatial analysis utilizing either drones, UAVs, or satellite data can be conducted, efficiently mitigating considerable mosquito transmission. Using CNNs for efficient mosquito habitat identification proves the most suitable method in disease and epidemic control, enabling governments and epidemiologists to identify areas that are vulnerable without costly and time inefficient approaches such as drones, lidar, and satellite data.

---

[1] The altitude of 1300 meters was utilized as a constant for satellite imagery in order to ensure consistency across training data. This altitude enabled large scale mosquito habitats to be detected with accuracy.

### 3.2 Comparison Against Existing Neural Networks

YOLO was utilized instead of other approaches such as a Faster RCNN due to its quicker speed resulting from a more optimal architecture. YOLO and Faster RCNN both employ bounding box regression, but YOLO performs classification and bounding box regression concurrently (Bochkovskiy *et al*, 2020), demonstrating a clear advantage in terms of run speed and efficiency. RCNNs additionally do not perform as well for real-time object detection (Dwivedi, 2020). They are better suited for small-scale object detection as they have nine anchors in a single grid compared to YOLO which has two anchors to predict a single class of objects. Since this paper deals with mosquito habitats on a macro scale, the YOLO algorithm was selected as it was more optimal for large-scale object detection where images did not have rare aspect ratios. In common classifying approaches mentioned in 3.1 such as drones, Faster RCNNs may prove more accurate due to the finer spatial scale for which the images are being analyzed for. However, due to this paper being a proof of concept, using YOLO, a faster network and having a model architecture better suited for macro-scale objects, proved to be the more effective solution.

Additionally, since this paper presents a proof of concept, the YOLO model was further used to demonstrate the effectiveness and viability of neural network models in comparison to other main approaches (i.e drones, UAVs, and satellite data), thus emphasis was not given towards other more difficult to implement models such as the RCNN.

### 3.3 Convolutional Neural Networks and Architecture/ Frameworks

#### 3.3.1 YOLOv4

YOLOv4, designed and implemented by Bochkovskiy *et al* (2020), is a single-stage convolutional neural network model. Prioritizing real-time detection, YOLOv4 operates at efficient speeds for inference. The network serves to predict object location and classification, utilizing bounding boxes to visually identify the objects desired. YOLOv4 utilizes CSPDarknet53 as a backbone, comprised of 27.6 million parameters. Using DarkNet-53, the backbone integrates CSPNet, introduced by Wang *et al* (2019), mitigating the requirement of heavy inference computations. An SPP (Spatial Pyramid Pooling) block (He *et al*, 2014) is added subsequently after CSPDarknet53, improving feature recognition and the receptive field. Employing PANet (Path Aggression Network), proposed by Liu *et al* (2018), for the model neck, YOLOv4 further aggregates information through better propagation of layer information, improving accuracy. For detection, YOLOv4 employs the model head from YOLOv3, implementing anchor-based detection. To improve the design further, Bochkovskiy *et al* implement new features such as Mosaic data augmentation, Mish activation, CIoU loss, DropBlock regularization, and more. Able to be trained with a single conventional GPU, YOLOv4 is optimized for simple configuration. YOLOv4 has been utilized and applied for a variety of purposes, such as melanoma lesion detection (Albahli *et al*, 2020), robotics (Gao *et al*, 2021), and surveillance (Kumar *et al*, 2020). Often applied for real-time detection and constant surveillance, its implementation in the domain of mosquito habitat surveillance and detection proves advantageous for active mosquito disease prevention efforts.

Figure 1: YOLOv4 Darknet Framework & Model Architecture

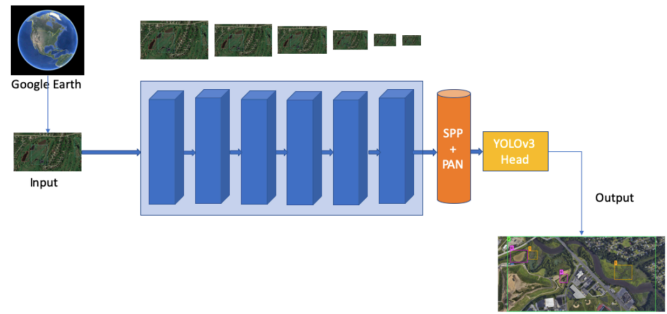

(©Google Earth 2021, Landsat/Copernicus Data SIO, NOAA, U.S. Navy, NGA, GEBCO)

#### 3.3.2 YOLOv5

YOLOv5 is a recent release of a family of YOLO models (Jocher, 2020). It is a single-stage object detector trained on the COCO dataset, combining bounding box prediction and object classification into a single differentiable framework. It includes a simple functionality for Test Time Augmentation (TTA), model ensembling, hyperparameter evolution, and export to ONNX, CoreML and TFLite.

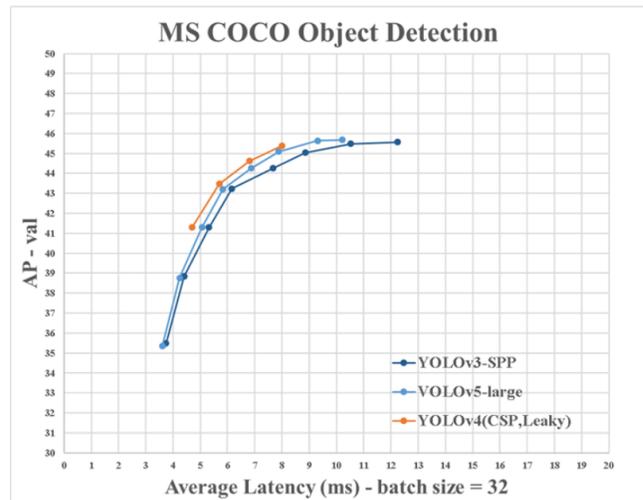

Figure 2: YOLOv4 Darknet Framework & Model Architecture

Compared to the Darknet framework of its counterpart YOLOv4, YOLOv5 utilizes the Pytorch framework which is very fast, lightweight, and user friendly. Like other single-stage object detectors, it consists of 3 important parts: Model Backbone, Model Neck, Model Head. YOLOv5 uses CSP

(Cross Stage Partial Networks) as the model backbone to extract important features from a given image. The model neck is used to generate feature pyramids for object scaling to identify the same object on different image orientations, sizes, and scales. In YOLOv5, PANet is used as a neck to get feature pyramids. The model head is used for the final detection part where it employs the Binary Cross-Entropy with Logits Loss function from PyTorch to generate final output vectors with objectness scores, class probabilities, and bounding boxes. Overall the model contains 191 layers, 7.46816e+06 parameters, and 7.46816e+06 gradients.

### 3.3.2.1 Activation, Optimization, and Loss Functions for YOLOv5

An activation function mimics the stimulation of a biological neuron, defining how the weighted input sum transforms into an output from nodes in layers of the network. It helps the network learn complex patterns by determining whether a neuron should be activated or not. This function acts as a gate between the input in the current neuron to the output going to the subsequent layer. YOLOv5 employs the Leaky RELU function (2) in the middle layers and the sigmoid function (1) in the final detection layer.

$$S(x) = \frac{1}{1 + e^{-x}} \quad (1)$$

LeakyRELU (2) is a self-regulated nonmonotonic activation function without zero slope parts and a mean activation approximate to zero, making training substantially faster and obtaining a more optimal result.

Mathematically, it is defined as follows (Maas *et al*, 2013)

$$f(x) = \begin{cases} 0.01x & \text{for} \quad x < 0 \\ x & \text{for} \quad x \geq 0 \end{cases} \quad (2)$$

The purpose of optimization functions is to find the model parameters that correspond to the best fit between actual and predicted outputs (Stojiljković, 2021). YOLOv5 utilizes a stochastic gradient descent optimization function which works by performing a parameter update for each training input, reducing redundancy. A stochastic gradient descent calculates the gradient using a small, random sample instead of the dataset as a whole. Each random set is called a minibatch and once all minibatches are used, the iteration (epoch) is finished and the subsequent one begins. The YOLOv5 model used in this study is run for 4500 epochs. The cost function is the function to be minimized (or maximized) by varying decision variables. In a convolutional neural network, weights and biases are used to minimize the difference between actual and predicted outputs by adjusting model parameters (Stojiljković, 2021). YOLOv5 calculates this compound loss based on the objectness score, class probability score, and bounding box regression score (Rajput, 2020). This model uses the Binary Cross-Entropy with Logits Loss function (3) which combines all operations into one layer, making it numerically stable.

$$H_p(q) = -\frac{1}{N} \sum_{i=1}^{N} y_i \cdot \log\left(p\left(y_i\right)\right) + (1 - y_i) \cdot \log\left(1 - p\left(y_i\right)\right) \quad (3)$$

### 3.3.3 YOLOR

YOLOR is a multimodal unified model that employs both explicit and implicit learning to accomplish various tasks. It uses the scaled YOLOv4 for explicit knowledge and employs implicit deep learning through Manifold Space Reduction, Kernel Alignment, and other functions such as offset refinement, anchor refinement, and feature selection. Similar to many CNN architectures, YOLOR uses the Backpropagation algorithm. Homogeneous to YOLOv5, it utilizes the PyTorch framework and is able to perform tasks of object detection, multilevel image classifications, and feature embedding. In this study, it performed with an increased speed of 88% compared to that of YOLOv4.

## 4. RESULTS AND DISCUSSION

### 4.1 YOLOv4 Image Classification Results and Analysis

Validation results of YOLOv4 are summarized in Table I, comprised of the metrics of Precision, Recall, and mAP. YOLOv4 was trained for 6000 epochs.

Table I: YOLOv4 Validation Data

| Images | Targets | Precision | Recall | mAP @0.5 IoU |
|---|---|---|---|---|
| 70 | 182 | 0.70 | 0.39 | 0.278248 |

Figure 3 presents the average loss against the number of epochs/iterations. The converging trend line indicates a positive learning rate where the model was neither under or overfitted. Seen in Figure 3, the loss correspondingly diminished as epochs increased, signifying successful learning of all classes as the loss approached 0 with increasing epochs. Loss rapidly decreases within the domain of 0 - 1200 epochs, indicating rapid learning of trends when first presented with training data (characteristic of CNNs), with a plateau of 0.5 as epochs increase.

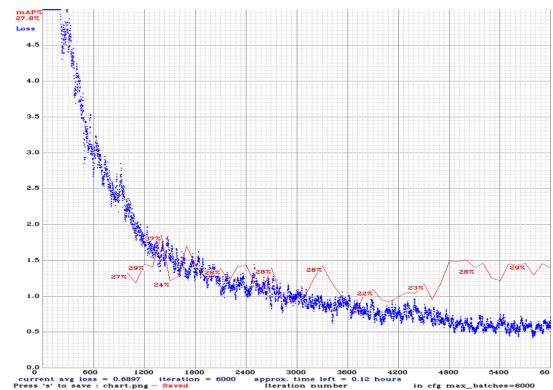

Figure 3: YOLOv4 Average Loss vs. Number of Epochs

Utilizing the trained YOLOv4 network on sample satellite imagery comprised of all classes (see Figure 4), it is observed the network accurately identifies all classes, with 1 river inlet upstream not identified (see 4.4 for explanation). All lakes/ponds (Class 0) were correctly identified, with a bounding box covering the entire extent of the water body surface. River inlets (Class 1) were identified accurately, missing 2 as a result of insufficient data of Class 1 for training (see 4.4). The river (Class 2) pictured in Figure 4 was accurately identified by the network. From the perspective of all satellite imagery inferences (see Figure 11), YOLOv4 is proven to be an ample network for mosquito habitat detection and a viable method in the application of mosquito and mosquito-borne disease surveillance.

Figure 4: YOLOv4 Inference Image Prediction

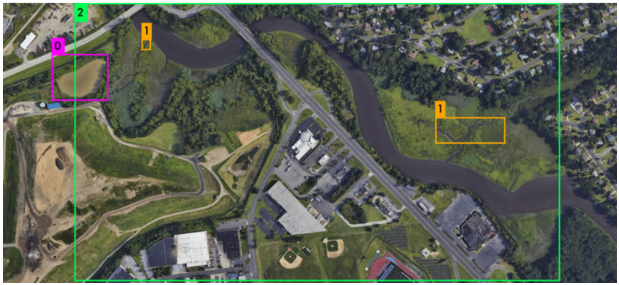

(©Google Earth 2021, Landsat/Copernicus Data SIO, NOAA, U.S. Navy, NGA, GEBCO)

*4.2 YOLOv5 Image Classification Results and Analysis*

The validation data for the YOLOv5 network is presented in Table II, consisting of Precision, Recall, and mAP. YOLOv5 was trained for 4500 epochs.

Table II: YOLOv5 Validation Data

| Images | Targets | Precision | Recall | mAP @0.5 IoU |
|---|---|---|---|---|
| 70 | 182 | 0.461 | 0.279 | 0.197 |

Trends, summarized in Figure 5, indicated successful network learning of classes and identification, often accompanied with an unstable beginning (see Precision, Recall, mAP, and val Classification). At approximately 3800 epochs, all trends appear to be stable and consistent, however, upon testing, consistency and accuracy diminished (see Figure 11 and Figure 6).

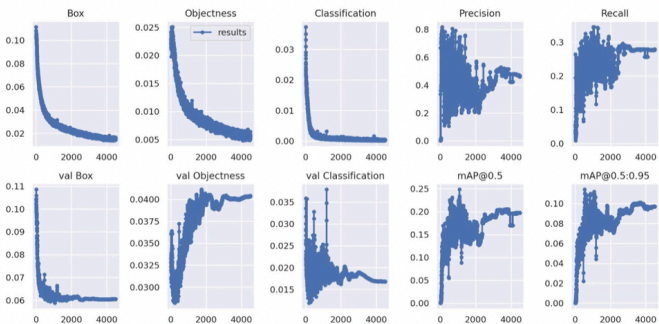

Figure 5: Validation plot diagrams summarizing YOLOv5 network training over epochs

The YOLOv5 model does not present an accurate and reliable model for the prediction of mosquito habitats (see Figure 6). As shown in Figure 6, the model predicted a solar panel as a pond/lake (Class 0) with a 74% confidence rate.

The inlet shown in the upper left corner was presented with a 51% confidence rate and wrongly classified as a pond/lake. It was unable to identify the pond/lake present in the upper left corner as well as 2 river inlets. With respect to all images (see further in Figure 11), YOLOv5 proves to be inconsistent with its mosquito habitat classification and therefore unreliable as an autonomous mosquito habitat detector.

Figure 6: YOLOv5 Inference Image Prediction

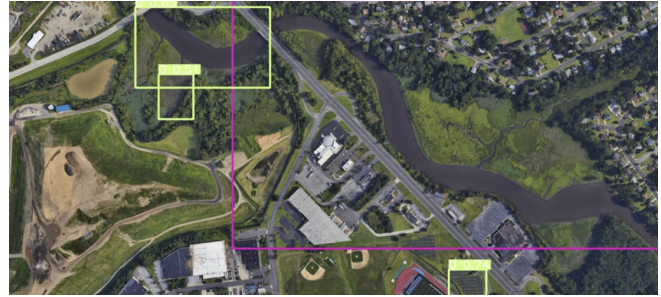

(©Google Earth 2021, Landsat/Copernicus Data SIO, NOAA, U.S. Navy, NGA, GEBCO)

*4.3 YOLOR Image Classification Results and Analysis*

Data results obtained from validation data, summarized in Table III, detail accuracy in mAP, Precision, and Recall. YOLOR was trained for 300 epochs with a batch size of 8 for training.

Table III: YOLOR Validation Data

| Images | Targets | Precision | Recall | mAP @0.5 IoU |
|---|---|---|---|---|
| 70 | 182 | 0.337 | 0.295 | 0.264 |

Training and validation accuracy is plotted within Figure 7, demonstrating instability in validation classification but characteristic improvements in other actions. Upon testing, however, the accuracy diminished substantiality, exhibited in Figure 8 and Figure 11.

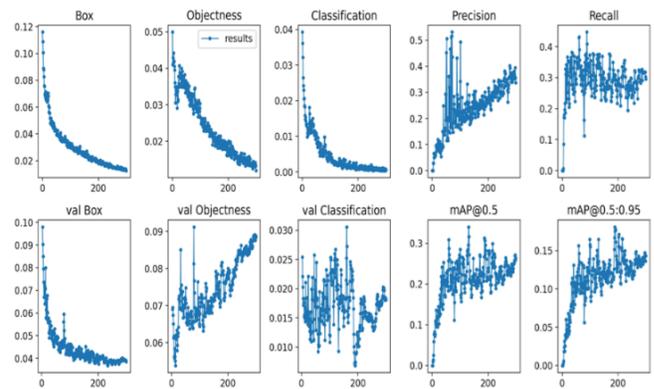

Figure 7: Validation plot diagrams summarizing YOLOR network training over epochs

Testing the trained YOLOR network on the utilized constant image composed of all three classes (see Figure 8), it is evident that the model is insufficient and unreliable in large-scale inference/detection of mosquito habitats. Within Figure 8, YOLOR is unable to identify the entire river (Class

3). In inference for river inlets (Class 2), it is able to identify smaller segments of the main inlet (East of the river) however is unable to identify the main inlet. It was able to identify 1 of 4 main inlets entirely. YOLOR failed in inference for Class 0 (Ponds/Lakes), identifying 0 ponds/lakes and further misclassifying other geographical characteristics as ponds/lakes, diminishing the effectiveness of YOLOR.

Figure 8: YOLOR Inference Image Prediction

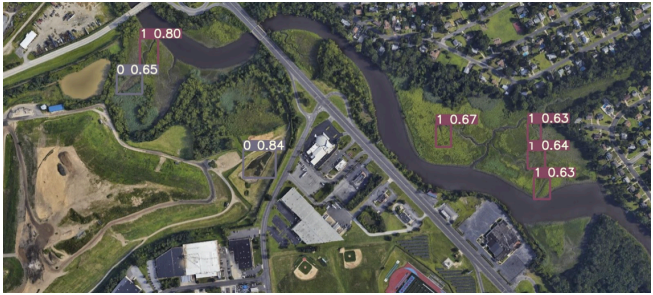

(©Google Earth 2021, Landsat/Copernicus Data SIO, NOAA, U.S. Navy, NGA, GEBCO)

*4.4 Convolutional Neural Network Model Comparison*

Figure 9: False Negative Satellite Imagery

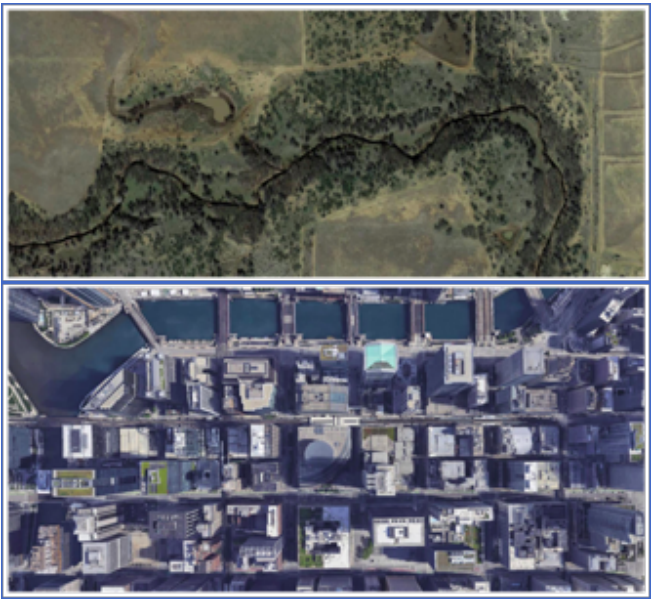

(©Google Earth 2021, Landsat/Copernicus Data SIO, NOAA, U.S. Navy, NGA, GEBCO)

The satellite imagery, seen in Figure 9, proved challenging for all CNN models to interpret, consistently outputting a false negative result. It can be assumed, based on the training data provided, that the model was not sufficiently trained on such terrain. This result can be prevented with a larger, and thus more diverse dataset, enabling the network to scale to different environments and extremities in terrain for mosquito habitat identification.

Figure 10: CNN Inference on Mountainous Terrain

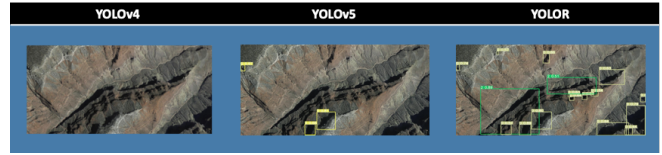

(©Google Earth 2021, Landsat/Copernicus)

Seen in Figure 10, there were extraneous cases where shadows were misclassified as possible mosquito habitats. Although training data was insufficient to enable accurate prediction within mountainous terrains[2], YOLOv4 was nevertheless able to identify that no mosquito habitat existed within the imagery based on prior trends learned.

Figure 11: Satellite Imagery Inference CNN Comparison

| Model | YOLOv4 | YOLOv5 | YOLOR |
|---|---|---|---|
| 1 | | | |
| 2 | | | |
| 3 | | | |
| 4 | | | |
| 5 | | | |

(©Google Earth 2021, Landsat/Copernicus Data SIO, NOAA, U.S. Navy, NGA, GEBCO)

The output/inference image sample comparison between the three tested models is presented in Figure 11. For ponds/lakes (Class 0), there is an overall higher confidence level across all 3 models. It is assumed that the training data is directly correlated with this observation, as there was an abundance/majority of training data dedicated for Style 0. The confidence levels were correspondingly high when the satellite image had a sharp contrast in landscape and defined features in contrast to the mosquito habitats. Evidence of this correlation is seen best in Figure 11 (3), where all 3 models accurately predicted the pond/lakes in the golf course landscape. This result may be attributed to its prominent appearance in contrast to the grass surrounding it. Inversely, this effect/correlation is identified in Figure 9 (1), where there existed a minimal contrast in landscape/color, preventing an accurate inference on the imagery consistent with all models.

Across all testing satellite imagery, excerpt shown in Figure 11, YOLOv4 performed more accurately relative to YOLOR & YOLOv5, outperforming in all classes and environments. Integral & substantial differences existed in terms of accuracy between the outputs of YOLOv4 and YOLOR/YOLOv5. Seen in Figure 11 (1), both YOLOR & YOLOv5 infer solar panels as ponds/lakes with > 50% confidence levels; comparatively YOLOv4 accurately determined that the solar panels did not belong within any trained class. Applying IoU (Intersection over Union) for relative comparison, YOLOv4 obtained a more optimal score. Consistent on all satellite imagery in Figure 11, YOLOv4 accurately determined dimensions and class of mosquito habitats. Important to note, YOLOv4 was time-intensive relative to the other models, though minimal enough for large-scale inference and application; further, it yielded optimal and conclusive results. Through the addition of more classes and training data, including dense vegetation, tall grasses, and other viable large-scale mosquito habitats, the proposed approach can be expanded to reach diverse topographical regions and reflect potential hotspot concentrations across the world.

---

[2] Suburban & Rural Flatlands/Forests comprised the majority of the training dataset

## 5. Conclusion

Compared to other aerial approaches for mosquito habitat detection, Convolutional Neural Networks(CNN) are the most efficient and cost-effective approach. CNNs provide an optimal alternative for macro-scale mosquito habitat detection that can be applied to identify potential hotspot regions of mosquito-borne diseases while multispectral imaging technology is better suited for micro-scale detection. The dataset was constructed using readily available Google Earth satellite imagery and annotated using bounding boxes for object detection on 3 different classes: Ponds/Lakes, Rivers, and River Inlets. This dataset was utilized to examine all 3 CNN models (YOLOv4, YOLOv5, YOLOR). YOLOv4 performed with the highest accuracy amongst the 3 models with an average IoU score of 55.85%. The image inferences demonstrate the model's high sensitivity and capability to correctly classify mosquito habitats in satellite imagery. Micro-scale habitats such as puddles, footprints, and tires go unnoticed by this model as its scope is geared towards larger bodies of standing water. It also does not factor in nearby vegetation and weather-related data that may have an impact on breeding patterns and female oviposition. Future research could include an integration of such factors mentioned to more accurately define viable mosquito habitats. The dataset could further be expanded to present a variety of different topographical regions to expand the influence and accuracy of the model in regions with distinct physical features and geography. In addition to expanding the dataset, more classes can be incorporated such as dense forest clusters and tall grasses to account for varying breeding sites across numerous mosquito species. The proposed solution can aid in preventative measures against the global transmission of mosquito-borne vector diseases through risk mapping as well as integration into public health policies. A convolutional neural network approach is additionally able to map impoverished and hard-to-reach areas, assisting vulnerable regions in mosquito-borne disease control. In scenarios such as mosquito-borne disease epidemics, it is optimal to utilize CNNs for immediate identification of large-scale mosquito habitats. During the treatment process of said habitats, a finer, more thorough analysis can be conducted using methods such as drones, UAVs, and satellite data. Through this approach, epidemics can be controlled efficiently and at a rapid pace by locating and treating potential hazardous mosquito habitats. There are a multitude of cases where efficient mosquito habitat detection and surveillance are needed for public health control; CNNs prove as a viable, cost-effective, and autonomous approach to achieving this.

## 6. Data/Code

Data and code to replicate the results of this experiment are available on a public Github repo: https://github.com/sriramelango/CNN-Mosquito-Detection

## 7. Acknowledgements

We would like to thank the NASA SEES Internship program for giving us the background knowledge and support needed to start this project. We also thank our mentors Russane Low, Peder Nelson, and Cassie Soeffing for their tremendous support and guidance in the research and publication process. We would like to thank Roboflow for giving us the software capabilities of organizing our dataset and importing them into Google Colab Notebook.

This publication has been possible thanks to the following authorization, grants, and permits. GLOBE is an interagency program funded by the National Aeronautics and Space Administration (NASA) and the National Science Foundation (NSF), supported by the U.S. Department of State, and implemented through a cooperative agreement between NASA and the University Corporation for Atmospheric Research (UCAR) in Boulder, Colorado. The GLOBE Observer app and SEfES Earth Explorers team are supported through a National Aeronautics and Space Administration (NASA) cooperative agreement: NNX16AE28A to the Institute for Global Environmental Strategies (IGES) for the NASA Earth Science Education Collaborative (NESEC). The SEES High School Summer Intern Program is led by the Texas Space Grant Consortium at the University of Texas at Austin (NASA Award NNX16AB89A).

**AUTHOR INFORMATION**

**Sriram Elango,** Researcher, Center for Space Research, University of Texas at Austin, Austin, TX

**Nandini Ramachandran,** Researcher, Center for Space Research, University of Texas at Austin, Austin, TX

**Russane Low,** Senior Scientist, Institute of Global Environmental Strategies, Arlington, VA